\documentclass[twocolumn,conference]{IEEEtran}
\usepackage[T1]{fontenc}
\usepackage[latin9]{inputenc}
\usepackage{array}
\usepackage{float}
\usepackage{calc}
\usepackage{textcomp}
\usepackage{multirow}
\usepackage{amsmath}
\usepackage{amssymb}
\usepackage{graphicx}
\usepackage{subscript}
\usepackage{color}

\makeatletter

\providecommand{\tabularnewline}{\\}

\usepackage[caption=false,font=footnotesize]{subfig}

\makeatother

\begin{document}

\title{Evaluating Graph Signal Processing for Neuroimaging Through Classification and Dimensionality Reduction}

\author{\IEEEauthorblockN{Mathilde~M\'enoret, Nicolas~Farrugia, Bastien~Pasdeloup and Vincent~Gripon}\IEEEauthorblockA{IMT Atlantique\\
Electronics Department\\
29200 Brest, France\\
Email: firstname.lastname@imt-atlantique.fr}}


\IEEEaftertitletext{after title text like dedication}
\maketitle
\begin{abstract} 
Graph Signal Processing (GSP) is a promising framework to analyze multi-dimensional neuroimaging datasets, while taking into account both the spatial and functional dependencies between brain signals. In the present work, we apply dimensionality reduction techniques based on graph representations of the brain to decode brain activity from real and simulated fMRI datasets. We introduce seven graphs obtained from a) geometric structure and/or b) functional connectivity between brain areas at rest, and compare them when performing dimension reduction for classification. We show that mixed graphs using both a) and b) offer the best performance. We also show that graph sampling methods perform better than classical dimension reduction including Principal Component Analysis (PCA) and Independent Component Analysis (ICA).
\end{abstract}


\IEEEpeerreviewmaketitle{}

\section{Introduction}

Analyzing neuroimaging data is a major challenge due to several intrinsic limitations of neuroimaging datasets (i.e. high sensitivity to noise, large number of dimensions for few observations per subject, etc.). While many discoveries in neuroscience have been made using massively univariate statistics, there has been a recent paradigm shift towards the application of multivariate analysis and machine learning to ``decode'' brain functions~\cite{Varoquaux2014}. The relevance of considering multivariate dependencies in brain signals is further justified by the rapidly growing literature on the application of network science and graph theory for studying brain connectivity~\cite{sporns2017future}. Surprisingly, few analysis methods take into account both the multivariate aspect and connectivity features of the brain, such as structural connectivity (white matter tracts), functional connectivity (i.e. statistical dependencies between signals over time) or simply geometrical relationships between observations. 

A promising avenue to address this important gap resides in Graph Signal Processing (GSP)~\cite{shuman_emerging_2013}. GSP is an emerging subfield of signal processing whose objective is to take into account the underlying graphical structure of multivariate data, in order to generalize common signal processing techniques (such as filtering, deconvolution, denoising, or time-frequency analysis) to irregular graph/network domains. GSP is built on the idea that the eigenvectors of the graph Laplacian matrix are analogous to Fourier modes, and can thus be used to provide a spectral representation of signals defined on a graph, through the so-called Graph Fourier Transform operator (GFT). In this paper, we evaluate the application of GSP for the analysis of neuroimaging data. More specifically, we assess whether GSP can lead to more accurate supervised classification, as well as whether GSP can be used for dimensionality reduction. 

Methods for GSP are still under active research, with applications such as the analysis of temperature sensor data~\cite{girau2015} or epidemiology~\cite{Segarra2017}. Also, GSP-based methods have recently been applied to neuroimaging using fMRI~\cite{huang_graph_2016} and EEG/MEG data~\cite{rui_dimensionality_2016,graichen_sphara_2015,Smith_2017}. 
Huang and collaborators~\cite{huang_graph_2016} have applied graph frequency analysis to fMRI data in order to observe how brain activity changes during a learning task. Graph frequency analysis allows to study spatial variation of the signal, with low graph frequencies representing smooth and regular variations across the brain network, whereas high graph frequencies represent important spatial variations, described by the authors as randomness. After decomposing fMRI data into graph frequency bands, Huang et al. observed that during learning, low graph frequencies correlate with the learning rate at the start of the training, while higher frequencies correlate with participants' familiarity with the task. 

Other studies have applied GSP techniques to EEG/MEG signals, for instance for noise suppression~\cite{graichen_sphara_2015}, dimensionality reduction~\cite{rui_dimensionality_2016,graichen_sphara_2015,liu2016simultaneous} and classification~\cite{rui_dimensionality_2016,liu2016simultaneous}.
The authors of the latter article compared classification accuracy when building graphs using different connectivity measures. They showed that projecting the data into the eigenspace associated with the graph strongest eigenvalues in order to reduce dimensions leads to better classification results than Principal Component Analysis (PCA) and Linear Discriminant Analysis (LDA)~\cite{rui_dimensionality_2016}. While these studies provide promising results suggesting a positive impact of GSP to EEG/MEG analysis, an important drawback is the lack of geometrical information in the construction of the graph edges. 

In the present paper, we aim to evaluate whether GSP can positively impact classification after dimensionality reduction in functional MRI (fMRI) datasets. We propose to study the influence of different types of graphs on brain signal classification, taking into account either geometrical or statistical dependencies between voxels, or both. To do so, we use the Graph Fourier Transform (GFT) to decompose brain signals into spectral components, compare several methods for dimensionality reduction of the decomposed signals (namely, graph sampling or statistical selection), and compare the performance of these methods to state-of-the-art reduction techniques such as PCA and ICA. We perform our experiments on two datasets, a simulated fMRI dataset and a real open source fMRI dataset~\cite{haxby_distributed_2001}. 

\section{Methods}

\subsection{Graph Signal Processing}
Throughout this paper, we consider a weighted graph $\mathcal{G}$, consisting of a set $\mathcal{V}$ of vertices indexed from $1$ to $N$ ($\mathcal{V} = \{v_1,\dots,v_N\}$), and of an adjacency matrix ${\bf W}$, such that ${\bf W}_{ij}~\in~\mathbb{R}^+$ denotes the weight between vertices $v_i$ and $v_j$. We consider symmetric ($\forall i,j : {\bf W}_{ij}={\bf W}_{ji}$) graphs. In the following sections, we introduce several graphs built from geometrical and/or statistical properties of the fMRI signals.

The Laplacian matrix of a graph is defined by ${\bf L} = {\bf D} - {\bf W}$, where ${\bf D}$ is the diagonal matrix of degrees defined by $\forall i: {\bf D}_{ii} = \sum_j {\bf W}_{ij}$. Being symmetric and real-valued, ${\bf L}$ can be decomposed as ${\bf L} = {\bf F} \boldsymbol{\Lambda} {\bf F}^\top$, where ${\bf F}$ is an orthonormal matrix, ${\bf F}^\top$ is its transposed matrix, and $\boldsymbol{\Lambda}$ is the diagonal matrix of eigenvalues, in ascending order.

A signal over $\mathcal{G}$ is a vector $\mathbf{x} \in \mathbb{R}^N$ interpreted as scalars observed on each vertex. Its Graph Fourier Transform (GFT) is given by $\hat{\mathbf{x}} = {\bf F}^\top \mathbf{x}$. The first coordinates of $\hat{\mathbf{x}}$, associated with the lower eigenvalues in $\boldsymbol{\Lambda}$, are called \emph{low frequencies} (LF) and its last are called \emph{high frequencies} (HF).

As far as our application case is concerned, vertices correspond to regions of interest in the brain. We denote by $\mathbf{X}$ a matrix of all measures obtained during rest periods (containing $M$ columns and $N$ lines) obtained from all subjects. These measures are distinct from the ones we aim to classify, and serve as a baseline that incorporate average statistical dependencies which are used to build the graphs. We denote by $\mathbf{X}_{:m}\in\mathbb{R}^N$ the $m$-th observation (column) and by $\mathbf{X}_{i:}\in\mathbb{R}^M$ the $i$-th row of $\mathbf{X}$, corresponding to all measures at vertex $v_i$.

\subsection{fMRI dataset}
We use an fMRI data provided openly by Haxby et al. 2001~\cite{haxby_distributed_2001}. This dataset consists of fMRI scans of 6 subjects during a visual stimulation experiment. For each subject, 1452 volumes of size $40\times 64\times 64$ (voxel size $3.5\times 3.75\times 3.75$ mm) were recorded every 2.5 seconds. The experiment is a block design with 12 sessions in which 8 types of stimuli (human faces, houses, cats, chairs, scissors, shoes, bottles and scrambled images) were presented during blocks of 24 seconds separated by 12 seconds of rest. Further details on the experiment are described in~\cite{haxby_distributed_2001}. Volumes were normalized in MNI space. We restrict our analysis to two contrasts: Face vs House and Cat vs Face.

\subsection{Simulated fMRI data}
We simulate fMRI datasets of size $53\times 63\times 46\times 421$ (corresponding to 3 mm\textsuperscript{3 } isotropic voxels, and a volume repetition time of 2 seconds) with NeuRosim~\cite{welvaert_neurosim:_2011}, an R-software package. The activations of 6 areas are simulated depending on two conditions. The areas are modeled as spheres whose centers correspond to the MNI coordinates of brain areas known to be involved in visual processing. We use a baseline obtained from the Haxby dataset (the averaged data from the rest conditions of subject 2). The experimental design is a block design with 12 blocks of 22 seconds per condition, separated by 10 seconds of rest. A one minute rest period is also included at half the experiment. The haemodynamic response is simulated using the Balloon model with the parameters described in~\cite{Buxton04}. We simulate noise as a mixture of Rician system noise, temporal noise of order 1, spatial noise, low-frequency drift, physiological noise (due to heart and respiration rates) and task-related noise, as described in~\cite{welvaert_neurosim:_2011}.

A total of 86 ``subjects'' are simulated by randomly varying the coordinates of the spheres, the activation magnitude of each area, and the signal-to-noise ratio (from 1.4 to 4.8). We calculate 20 simulations for each ``subject'', resulting in 1720 simulations in total.  

\subsection{Data preprocessing}
Both fMRI datasets are analyzed with nilearn and scikit-learn~\cite{pedregosa_scikit-learn:_2011}. All data is normalized in MNI space and parcellated into 444 symmetrical regions of interest using the BASC atlas~\cite{bellec_multi-level_2010}. We compute the coordinates of the baricenters of the ROIs to obtain geometrical relationships between ROIs. The fMRI data are high-pass filtered at 0.01Hz, and no spatial smoothing is applied. The data used for classification is the raw BOLD signal of the 444 regions: for Haxby, 9 volumes in each block are used in the analysis resulting in 108 volumes per condition. For the simulated data, the first volume of each block is removed to account for the delayed haemodynamic response, therefore 10 volumes in each block are used, resulting in 120 volumes per condition.

\subsection{Graph construction}
We consider seven different graphs for each subject/simulation. Two graphs model the geometric structure of the brain (\emph{Full} and \emph{Geometric}) by setting weights using a Gaussian kernel (with empirically determined parameters) of the Euclidian distance between the barycenters of the 444 brain areas. The \emph{Full} graph is fully connected and the \emph{Geometric} only connects close brain areas (distance inferior to a radius, empirically determined), the weights of edges between distant brain areas being set to 0.

Three other graphs model the functional connectivity at rest between the brain areas, using different connectivity measures: absolute values for \emph{Correlation} and \emph{Covariance}, and the method by \emph{Kalofolias}~\cite{kalofolias2016learn} to infer a graph Laplacian matrix ${\bf L}$ from signals, assuming smoothness of the observed signals on the inferred graph.

Finally, two other graphs mix both the structure and connectivity of the brain: the \emph{Semilocal} graph connects only close brain areas (as the \emph{Geometric} graph) but  its weights correspond to the covariance between those brain areas. The \emph{Fundis} graph is defined in~\cite{rui_dimensionality_2016} as a product of distance and connectivity.

The following equations sum up these graphs, where $d\left(v_i,v_j\right)$ denotes the Euclidean distance between vertices $v_i$ and $v_j$ according to their spatial coordinates; $\sigma,\alpha,\beta$ and $\theta$ are empirically determined parameters; $\mathbf{X}$ is the matrix of all measurements; $\mathbf{Y}$ is an optimization parameter with same dimensions as $\mathbf{X}$; and $\mathcal{L}$ denotes the set of Laplacians:

\vspace{0.2cm}
\noindent
\textbf{Geometric graphs:}

\noindent
\emph{Full}:~~~~${\bf W}^{(full)}_{ij}=\exp\left(-\frac{d\left(v_i,v_j\right)^{2}}{2\sigma}\right)$

\noindent
\emph{Geometric}:~~~~${\bf W}^{(geo)}_{ij}=\begin{cases}
{\bf W}^{(full)}_{ij} & \text{if }\,d\left(v_i,v_j\right)<\alpha\\
0 & \text{otherwise}
\end{cases}$

\noindent
\textbf{Functional graphs:}

\noindent
\emph{Absolute correlation}:~~~~${\bf W}^{(corr)}_{ij} = | corr\left({\bf X}_{i:},{\bf X}_{j:}\right)|$

\noindent
\emph{Absolute covariance}:~~~~${\bf W}^{(cov)}_{ij} = | cov\left({\bf X}_{i:},{\bf X}_{j:}\right)|$

\noindent
\emph{Kalofolias}: \\
${\bf L}^{(kal)} = \arg\min\limits_{{\bf L}\in\mathcal{L}, {\bf Y}} \displaystyle{\sum_{m=1}^{M}{\| {\bf X}_{:m} - {\bf Y}_{:m} \|_2^2 + \beta ({\bf Y}_{:m})^\top {\bf L} {\bf Y}_{:m}}}$

\noindent
\textbf{Mixed graphs:}

\noindent
\emph{Semilocal}:~~~~${\bf W}^{(semi)}_{ij}=\begin{cases}
{\bf W}^{(cov)}_{ij} & \text{if }\,d\left(v_i,v_j\right)<\alpha\\
0 & \text{otherwise}
\end{cases}$

\noindent
\emph{Fundis}:~~~~${\bf W}^{(fun)}_{ij}=\exp\left(-\frac{\left(1-{\bf W}^{(corr)}_{ij}\right)^{2}}{2\theta}-\frac{d\left(v_i,v_j\right)^{2}}{2\sigma}\right)$
\vspace{0.2cm}

We use the eigenvectors of the Laplacian matrices of these graphs to perform GFT of the acquired signals, but also to reduce dimension using graph sampling methods, as explained in the following paragraphs. All GSP operations were done using the Matlab and Python versions of the GSP toolbox~\cite{perraudin2014gspbox}.

\subsection{Dimensionality reduction}
In our experiments, we consider different methods of dimensionality reduction. In particular we compare graph sampling (GS) to other graph-free methods: PCA, ICA, and selection of the $K$ best components using analysis of variance (ANOVA).

GS is a method adapted from~\cite{puy_random_2016} to select the vertices where the signal energy is the most concentrated. To do so, we compute the graph weighted coherence for a frequency band of interest delimited by indices ($f_{min}$, $f_{max}$), and extract the $K$ vertices achieving maximum scores. The graph weighted coherence for vertex $v_i$ is defined as $\sum_{k=f_{min}}^{f_{max}}{{\bf F}_{ik}^2}$. We restrict our analysis to either only \emph{low frequencies} (LF) (below $N/2$) of \emph{high frequencies} (HF) (above $N/2$). 

We also apply a method to perform graph frequencies sampling~\cite{rui_dimensionality_2016}, where we select $K$ eigenvalues (HF/LF/ANOVA), then project the signals to keep only the corresponding components. 
We present the results for $K=50$ components.

\subsection{Classification}
Classification is performed to disentangle brain signals originating from different conditions using different methods: linear Support Vector Machine (SVM), $k$-nearest neighbors ($k=15$) and logistic regression with $l_1$ penalty. To avoid excessive over-fitting and given the block design, cross-validation is performed across different sessions, leaving two sessions out: 16\% of the data is used as test data, the remaining as training. Classification is performed on the fMRI data (all data or reduced data: PCA/ICA/ANOVA/GS) and on the signal projected in the graph Fourier domain using GFT (reduced data HF, LF and ANOVA). After dimension reduction, data is standardized. All the procedure --- standardization, dimension reduction (for PCA, ICA and ANOVA when applied) and classification --- is cross-validated in a pipeline. We only detail SVM cross-validated accuracy scores due to lack of space. Results for other classifiers follow the same trend.

For the simulated fMRI data, classification is performed for each simulation. We average the results of 20 simulations per ``subject'', resulting in one value for each. The classification of the full brain fMRI data serves as reference.

\subsection{Statistics}
We perform a statistical analysis of accuracy scores across methods in order to estimate the significance of accuracy gain/loss. Non-parametric Friedman tests for repeated measures (an analysis of variance by ranks) are computed to identify differences between the conditions. Wilcoxon tests are used as post-hoc tests (Bonferonni adjusted for multiple comparisons). For the Haxby dataset, the same statistical tests are performed for exploratory purposes, however they should be interpreted with caution since the sample size is small. 

\section{Results}

We present classification results for the raw data using all 444 areas, to define a reference for further comparisons. For the Haxby dataset, Face vs. House classification achieves on average an accuracy of 88.4\% \textpm{} 4.4, and Cat vs. Face achieves 69.8\% \textpm{} 6.8. For the simulated data, accuracy ranges from 55.9\% to 95.9\%. In order to compare our results with the Haxby dataset, two groups of simulated ``subjects'' are determined : \emph{Easy} (accuracy > 80\%) comprised of 40 simulated subjects (average 89.8 \% \textpm{} 4.0), and \emph{Difficult} (accuracy between 55\% and 80\%) comprised of 46 simulated subjects (average 67.0 \% \textpm{} 6.4).


\subsection{Dimensionality reduction}
Several GSP-based methods for dimensionality reduction have been compared. Classification results for all methods are presented in Table \ref{tab:accuracy}.

\begin{table}[h]
\caption{Accuracy of the classification for all graph-based sampling methods
(in \%) for the two difficulty groups. Starred numbers indicate best scores in their category.}
\begin{center}
  \begin{tabular}{|c|c|c|c|c|c|}
    \hline 
    Graph & GFT  & GFT  & GFT & GS  & GS \tabularnewline
    Types & LF & HF & ANOVA & LF &  HF \tabularnewline
    \hline
		\multicolumn{6}{c}{Difficult}\tabularnewline  
    \hline
    Full & 54.8\%  & 51.1\%  & \textbf{66.0\%} & 52.0\% & 51.3\% \tabularnewline
    \hline 
    Geometric & 56.7\%  & 64.8\% & 64.8\% & 50.5\% & 65.2\%\tabularnewline
    \hline
    |Correlation| & 52.4\% & 66.8\%  & 64.7\% & 50.9\% & 60.3\% \tabularnewline
    \hline
    |Covariance| & 52.4\% & 67.6\%  & 65.2\% & 51.2\% & 66.2\% \tabularnewline
    \hline
    Kalofolias & \textbf{61.6\%}  & 51.9\%  & 65.9\% & \textbf{61.6\%} & 51.9\%\tabularnewline
    \hline
    Semilocal & 53.8\%  & \textbf{69.5\%}  & 65.6\% & 50.3\% & \textbf{72.5\%}* \tabularnewline
    \hline
    Fundis & 54.9\%  & 64.2\%  & 65.1\% & 49.7\% & 62.8\%\tabularnewline
    \hline
		\multicolumn{6}{c}{Easy}\tabularnewline 
    \hline
     Full & 65.1\%  & 60.0\%  & 88.9\% & 49.6\% & 60.6\%\tabularnewline    
    \hline 
    Geometric & 71.3\% & 79.4\%  & 86.0\%  & 57.8\%  & 79.1\% \tabularnewline
    \hline
    |Correlation| & 59.5\%  & 86.5\%  & 86.9\% & 53.5\%  & 75.2\% \tabularnewline
    \hline
    |Covariance| & 57.2\%  & 87.0\% & 88.8\% & 52.8\% & 84.8\%\tabularnewline
    \hline 
    Kalofolias & \textbf{88.2\%}  & 54.2\%  & \textbf{89.4\%} & \textbf{87.5\%} & 52.6\%\tabularnewline
    \hline 
    Semilocal & 61.3\% & \textbf{87.9\%} & 88.6\% & 52.3\% & \textbf{90.9\%}*\tabularnewline
    \hline
    Fundis & 67.4\%  & 77.5\%  & 86.7\% & 52.4\%  & 77.5\%\tabularnewline
    \hline 
  \end{tabular} 
\end{center}
\label{tab:accuracy}
\end{table}

\begin{figure}[tbh]
\centering\includegraphics[scale=0.5]{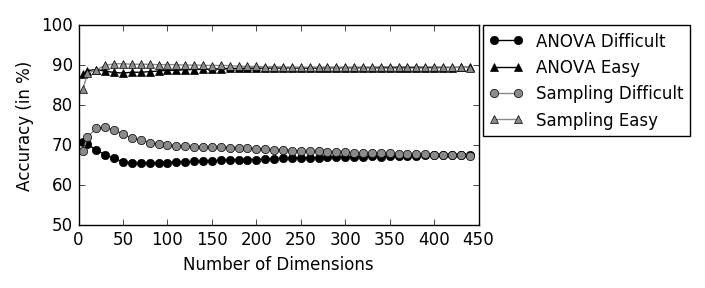}
\caption{Classification performance for the \emph{Semilocal} graph depending on the
number of dimensions. Comparison of the graph sampling method (gray)
and the graph K-best (black) for the two groups: \emph{Easy} and \emph{Difficult}.}
\label{fig:perfs}
\end{figure}

We observe that for all graph types but the \emph{Kalofolias} graph and the \emph{Full} graph, high frequencies are more relevant for the classification than low frequencies (classification is close to chance level in most cases). Moreover, the \emph{Semilocal} graph stands out from the other graph types and reaches better scores in both groups (72.5\%, 90.9\%). The \emph{Semilocal} graph was selected for further analysis. When comparing the optimal number of dimensions, GS yields the best accuracy when selecting 30 components for the \emph{Difficult} group and 50 components for the \emph{Easy} group (see Figure \ref{fig:perfs}).

\subsection{Comparisons of GSP, PCA, ICA and ANOVA}
The performance of GS to state-of-the-art reduction techniques such as PCA, ICA and ANOVA are then compared for simulated and real fMRI data. Table \ref{tab:gs} presents the results.

\begin{table}[h]
\caption{Comparison of Graph Sampling (\emph{Semilocal} graph), PCA, ICA and ANOVA. Classification accuracy
with 50 components for the Simulated and Haxby datasets.}
\begin{center}
  \begin{tabular}{|c|c|c|c|c|}
    \hline 
    \multirow{2}{*}{Method} & \multicolumn{2}{c|}{Simulation} & \multicolumn{2}{c|}{Haxby}\tabularnewline
    \cline{2-5} 
    & Easy & Difficult & Face-House & Cat-Face\tabularnewline
    \hline
    \hline
    PCA & 88.8\%  & 65.5\%  & 82.7\%  & 63.6\%\tabularnewline
    \hline 
    ICA & 90.2\%  & 65.3\% & 84.4\%  & 67.0\% \tabularnewline
    \hline 
    ANOVA & \textbf{92.1\%}  & 67.3\% & 85.5\%  & 65.5\%\tabularnewline
    \hline 
    \multirow{1}{*}{Graph sampling} & 90.9\% & \textbf{72.5\%} & \textbf{88.2\%} & \textbf{69.0\%}\tabularnewline
    \hline 
  \end{tabular}
\end{center}
\label{tab:gs}
\end{table}

For the simulated fMRI data, the classification with GS is significantly more accurate in the \emph{Difficult} group than PCA ($Z=5.9$, $p<0.001$), ICA ($Z=5.9$, $p<0.001$) and ANOVA ($Z=5.9$, $p<0.001$). In the \emph{Easy} group, classification is significantly more accurate for the ANOVA (PCA: $Z=4.4$, $p<0.001$; ICA: $Z=5.5$, $p<0.001$; and GS: $Z=3.7$, $p<0.001$). Classification with GS is significantly more accurate than PCA ($Z=5.2$, $p<0.001$), but not than ICA ($Z=2.0$, $p=0.05$).

For the Haxby dataset, the classification with GS produces the most accurate results for both conditions. However, those differences do not reach statistical significance (PCA: $Z=2.3$, $p=0.022$ uncorrected, ICA: $Z=1.5$, $p=0.126$ uncorrected, ANOVA: $Z=1.5$, $p=0.126$ uncorrected). 

\section{Conclusion}
In this work, we tested the contribution of Graph Signal Processing to brain signal analysis. We constructed graphs that model the geometric and/or the functional dependencies of brain activity on simulated and real fMRI data, and compared classification accuracy for difference choices of graphs and dimensionality reduction techniques. We showed that applying graph sampling to a semilocal graph could select meaningful vertices for classification, without any prior hypothesis on the categories to distinguish, and led to a significant improvement in classification accuracy compared to PCA, ICA and ANOVA when categories are difficult to distinguish. The semilocal graph best fits the data structure by taking into account both the geometric structure of the data and functional connectivity between brain areas at rest, and improves classification and dimensions reduction of neuroimaging data. 

We observe that LF features are better for the Kalofolias graph whereas
HF features are better for the semilocal graph. Those observations are expected since, for the
Kalofolias' method, by construction, the signals are smooth on the graph, i.e. all their energy is carried out by the first Fourier frequencies. For the semilocal graph on the opposite, while low graph frequencies might correspond to more stable activity spread across the brain for both categories (e.g. the gradient that develops from occipital to more frontal areas during visual processing), the discriminative features between considered conditions are localized in the brain, and thus carried out by high frequencies of geometric graphs. We believe this observation is particularly interesting for anyone interested in applying a similar methodology to perform classification: the components of signals to be kept are highly dependent on the method used to build the graph, as well as on the type of expected discriminative features between conditions.


To conclude, GSP is a promising method to improve the analysis of neuroimaging signals. Future work should focus on defining appropriate graphs, since it has a strong impact on the performance. Structural connectivity measures could provide additional information for graph construction.

\section*{Acknowlegment}
This work was funded by the Neural Communications project of the Labex CominLabs.

\bibliographystyle{IEEEtran}
\bibliography{Telecom_Graph}



\end{document}